\documentclass[runningheads]{llncs}
\usepackage[T1]{fontenc}
\usepackage{multirow}
\usepackage{graphicx}
\usepackage{tabularx}
\usepackage[usenames,dvipsnames]{xcolor}
\usepackage{amsmath}
\begin{document}
\title{ViTALS: \underline{Vi}sion \underline{T}ransformer for \underline{A}ction \underline{L}ocalization in  \underline{S}urgical Nephrectomy} 
\titlerunning{ViTALS}
%
\author{Soumyadeep Chandra\inst{1}\and 
        Sayeed Shafayet Chowdhury\inst{1}\and
        Courtney Yong\inst{2}\and
        Chandru P. Sundaram\inst{2}\and
        Kaushik Roy\inst{1}}
\authorrunning{Chandra et al.}
%
\institute{Elmore Family School of Electrical and Computer Engineering, Purdue University, West Lafayette, IN 47907, USA \and
School of Medicine, Indiana University, Indianapolis, IN 46202, USA\\
\email{\{chand133, chowdh23, kaushik\}@purdue.edu, {\{cyong, sundaram\}@iu.edu}}}
\maketitle              
\begin{abstract}
Surgical action localization is a challenging computer vision problem. While it has promising applications including automated training of surgery procedures, surgical workflow optimization, etc., appropriate model design is pivotal to accomplishing this task. Moreover, the lack of suitable medical datasets adds an additional layer of complexity. To that effect, we introduce a new complex dataset of nephrectomy surgeries called UroSlice. To perform the action localization from these videos, we propose a novel model termed as `ViTALS' (\underline{Vi}sion \underline{T}ransformer for \underline{A}ction \underline{L}ocalization in \underline{S}urgical Nephrectomy). Our model incorporates hierarchical dilated temporal convolution layers and inter-layer residual connections to capture the temporal correlations at finer as well as coarser granularities. The proposed approach achieves state-of-the-art performance on Cholec80 and UroSlice datasets (89.8\% and 66.1\% accuracy, respectively), validating its effectiveness.

\end{abstract}

\section{Introduction}
Action localization from surgical videos is a challenging video understanding task. Successful implementation of this task has several impacts such as (i) aiding in automated surgical training, (ii) serving as an introspective assistance tool, (iii) optimization of the surgical workflow, etc. With the proliferation of machine learning (ML) based computer vision applications, temporal action localization from videos has been studied widely for natural videos~\cite{czempiel2020tecno,yi2021asformer,zhang2022actionformer}. However, due to the scarcity of medical data (for privacy issues) and the difference in the overall scene variety and composition in different sub-tasks between natural and surgical videos, directly applying the existing approaches for surgical videos might not be optimal. On the other hand, surgical video understanding can strongly benefit from the strong modeling capabilities of today's deep learning models. To that end, there is a need to explore and innovate suitable ML models tailored for surgical videos.

Earlier studies on surgical action segmentation can be broadly categorized into single-stage or multi-stage models. Single-stage models often employ statistical approaches, utilizing surgical signal~\cite{blum2010modeling}, dynamic time-wrapping~\cite{padoy2012statistical}, and Hidden Markov Models (HMMs)~\cite{lalys2011surgical,padoy2008line} on visual features from input images. Bardham et al.~\cite{bardram2011phase} and Holden et al.~\cite{holden2014feasibility} explore sensor data for workflow segmentation, requiring additional manual annotations.

With the advent of deep learning, methodologies exclusively utilizing video frames have gained prominence. Pioneering works like EndoNet~\cite{twinanda2016endonet,twinanda2017endolstm} employ CNNs for recognition, while MTRCNet-CL~\cite{jin2020multi} adopts a multi-task framework for tool detection and phase recognition. SV-RCNet~\cite{jin2017sv} utilizes an end-to-end RNN (ResNet) architecture for frame extraction, followed by an LSTM network for capturing temporal dependencies within sequential frames. Yi et al.~\cite{yi2019hard} introduce Online Hard Frame Mapper (OHFM) to handle visually indistinguishable features, enhancing model robustness. OPERA~\cite{czempiel2021opera} introduces attention regularization, while Trans-SVNet~\cite{gao2021trans} proposes a hybrid spatio-temporal transformer aggregator to focus on high-quality frames during training. TeCNO~\cite{czempiel2020tecno} uses adaptive dilated convolution inspired by temporal convolution. AVT~\cite{girdhar2021avt} and ASFormer~\cite{yi2021asformer} leverage ViT-based backbones with causal dilated convolution for spatial and temporal feature extraction.

Drawing inspiration from existing works, our proposed model, `ViTALS,' is designed to enhance the multi-stage structure for surgical phase recognition tasks,  The absence of inductive bias in Vision Transformer (ViT) models makes it challenging to train effectively on small surgical datasets, leading to severe overfitting. To mitigate these issues, we introduce temporal convolution layers~\cite{ding2018weakly}, known for establishing a local connectivity inductive bias. This helps constrain the hypothetical space and facilitates effective learning of the target function with limited training sets. The hierarchical network, composed of encoder and decoder modules with an increasing dilation rate~\cite{zhu2020deformable}, enables learning appropriate weights for long sequential tasks. Lower encoder layers emphasize local features initially, gradually incorporating global dependencies at higher levels. Intermediary residual connections mitigate feature information loss from lower levels, and dilation enhances the scalability of transformers by improving time and memory complexity. In the decoder modules, a cross-attention mechanism preserves the initial prediction, enabling fine-tuning of smaller local features at each refinement stage. Experimental results on Cholec80~\cite{twinanda2016endonet} and UroSlice demonstrate the efficacy of our comprehensive approach in improving the model's performance in surgical phase recognition tasks.

\vspace{1em}
\noindent The specific contributions of this work:
\begin{enumerate}
    \vspace{-1em}
    \item We introduce a complex surgical dataset capturing right and left partial and radical nephrectomy surgeries. The phases in this dataset occur without uniformity and systematic order, introducing challenges due to the dataset's unpredictable nature and significant variations in temporal aspects.
    \item We proposed the model `ViTALS' to address challenges in surgical datasets by integrating temporal convolution layers with attention modules to counter inductive bias and refine temporal predictions by capturing local and global dependencies.
    \item ViTALS attains state-of-the-art results in surgical phase recognition, achieving 89.8\% accuracy on Cholec80 and 66.1\% on UroSlice datasets.
\end{enumerate}

\section{Methods}
This research tackles surgical video action segmentation by reframing the problem as a video phase classification task. For a video input stream $X$, each video frame $x_{i} \in {\mathbf{R}}^{H \times W \times C}$, represented as an RGB image (H x W x 3), is assigned a corresponding surgical phase label. A model denoted by $f_{\theta}$ aims to map each frame to its phase, resulting in an output $p_{t} \in {{k}}^K_{k=1}$ representing the probability distribution of the frame belonging to each of the $K$ possible phases within the specific surgical dataset.

\vspace{-1em}
\begin{figure}[!h]
\includegraphics[width=\textwidth]{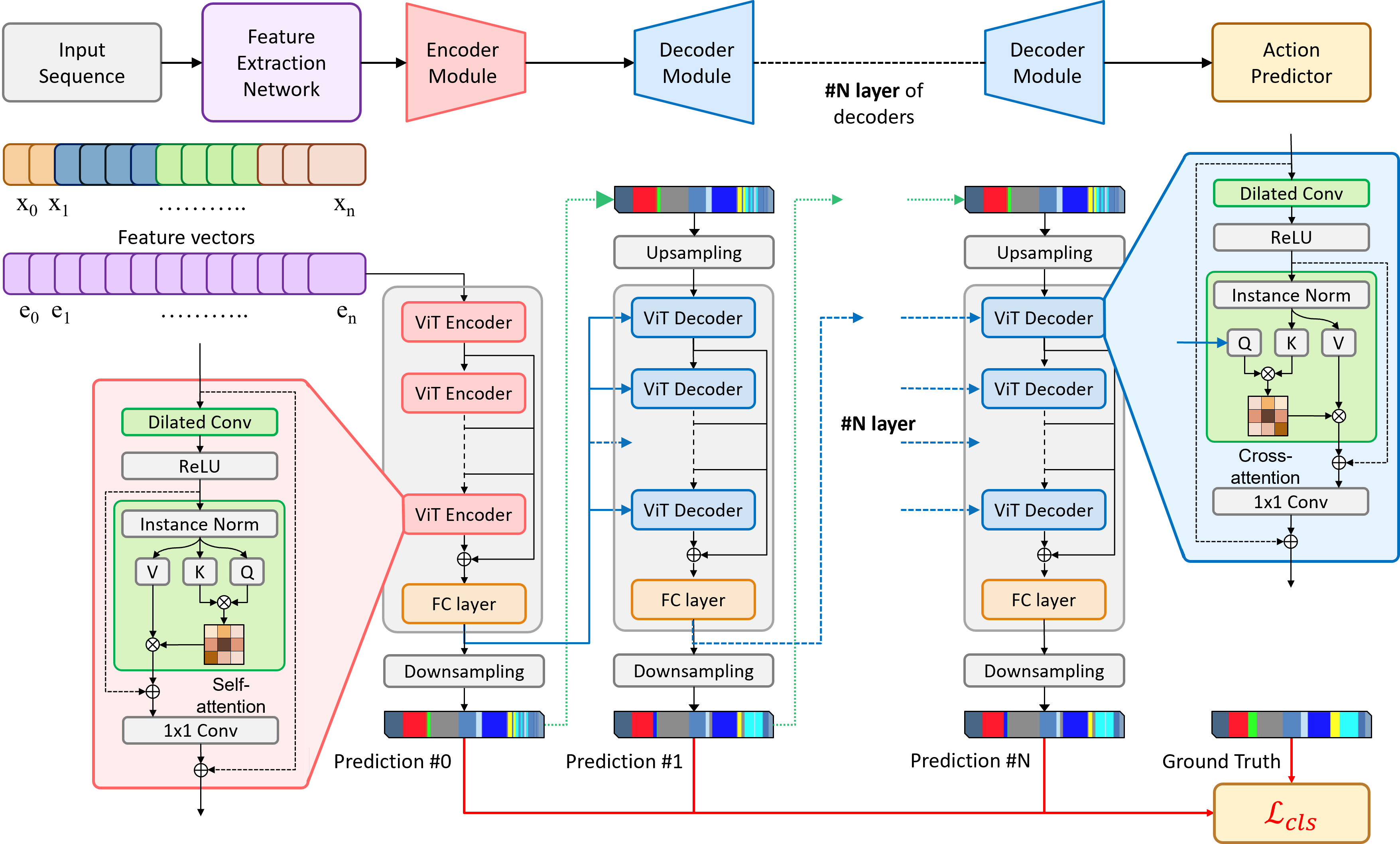}
\caption{Overview of our Proposed `ViTALS' model for surgical phase recognition. The feature extractor model extracts the spatial vector ($e_i$) information from the video sequence ($x_i$). The action segmentation network consists of multi-stage encoder-decoder(s) to perform refined phase prediction. The Encoder module (in \textcolor{red}{red}) consisting of a hierarchical network of individual ViT encoder blocks with intermediate residual connections produces an initial prediction. Within each of the encoder blocks, a dilated temporal feed-forward convolution layer is followed by a self-attention layer with residual connections. The Decoder module (in \textcolor{blue}{blue}) compromises a similar hierarchical network. However, in each decoder block the initial predictions are leveraged for a cross-attention mechanism to refine the granular predictions further.}  \label{fig:model}
\end{figure}

\vspace{-0.8cm}
\subsection{Model Overview} \label{model}
Inspired by ASFormer~\cite{yi2021asformer}, our `ViTALS' leverages a sophisticated approach, combining a temporally rich spatial feature extractor with a ViT-based encoder-decoder action segmentation network to address the intricate task of surgical action recognition as shown in Fig \ref{fig:model}. The input for the dataset consists of raw surgical video frames. To overcome GPU memory limitation, we use a pre-trained spatial feature extractor to embed long frame sequences and extract features from every video frame. These extracted features are then fed into the action segmentation network for recognizing surgical phases. In Section \ref{encoder}, we delve into how the encoder predicts the initial action probabilities across thousands of frames. These initial predictions serve as input to each decoder module to accurately refine the prediction and segment surgical actions in the video sequence.

\subsection{Spatial Feature extractor} \label{feature}
Surgical videos are often lengthy sequences with significant frame interdependence. However, their duration, ranging from a few minutes to several hours, poses a challenge for training prediction networks directly on raw video frames due to GPU limitations. To overcome this constraint, a separate feature extractor (FE) is trained with frame-level supervision to transform frame sequences into spatio-temporal feature vectors. The FE uses a kind of residual network called ResNet50~\cite{he2016deep_resnet} model trained independently for each surgery dataset with a 70-30 train-test split. Assuming that crucial temporal information is concentrated in the initial frames of each phase, we downsample the videos while retaining key temporal information for effective model training. Approximately 15,000 frames $X' = (x_{1}, x_{1+w}, x_{1+2w}, ..., x_n)$ spaced at equal intervals $w = \frac{n}{15000}$.

\vspace{-1em}
\begin{equation}
    \begin{split}
     E' &= \{e_{1}, e_{1+w}, ... e_n\} = FE(X') = \{FE(x_{1}), FE(x_{1+w}), ..., FE(x_n)\}
    \end{split}
\end{equation}

The trained feature extractor is further employed to generate feature vectors $E = \{{e_i}\}^{n}_{i=1}; \forall e_{i} \in {\mathbf{R}}^{d}$ for the entire extended sequence of video frames. The generated feature vectors are then seamlessly integrated into the subsequent stages of the proposed action segmentation network in `ViTALS'.

\subsection{ViTALS model}
\subsubsection{Encoder:} \label{encoder}
We designed an encoder-decoder architecture inspired by ViT, employing a single encoder and multiple decoder blocks. The video enters the model as spatial feature vectors ($E$, size: $n \times d$, where $n$ is video length and $d (2048)$ is feature dimension). The encoder, consisting of a series of smaller ViT encoders, processes this data. The encoder block of our architecture starts by analyzing local temporal features using a dilated temporal convolution. This technique is particularly effective for tasks like surgical video analysis, where limited data and highly correlated features exist within each action. Compared to point-wise fully connected layers, dilated convolutions introduce a `local inductive bias', leading to better performance~\cite{yi2021asformer}. The output then goes through a self-attention layer, where the interactions between layers become crucial for effective representation in surgical videos. Similar to findings in Temporal Convolution Networks (TCNs)~\cite{farha2019mstcn}, a hierarchical structure with increasing dilation rates allows learning local features while progressively capturing global information through an expanding receptive field. The dilation window size is augmented two-fold at each layer, denoted as $d_w = 2^i$; where $i=1,2,..L$. However, feature downsampling and sparse attention can diminish fine-grained information. To address this, a temporal fusion head integrates intermediate features from all layers to generate initial phase predictions $p_{0} \in {\mathbf{R}}^{n \times K}$ for each frame in the video (size: $n \times K$, where $K$ is the number of possible phases), enhancing the model's ability to capture both local and global context, vital for surgical video analysis.

\vspace{-1em}
\subsubsection{Decoder:} \label{decoder}
Earlier works have shown the application of additional TCN~\cite{farha2019mstcn} or GCN~\cite{huang2020improving_gcn} layers allow further refinement of the initial prediction by the primary network. The iterative refinement process of decoders ensures accurate and nuanced recognition of surgical phases as the information is processed through subsequent layers. The decoders exhibit a hierarchical architecture akin to the primary encoder block, incorporating cross-attention layers. This mechanism considers that the key (K) and value (V) are derived from the preceding layer, while the query (Q) originates from the previous encoder/decoder module. This separation of encoder and decoder~\cite{farha2019mstcn}, has been proven to improve the stability and effectiveness of the refinement process.

\vspace{-1em}
\subsubsection{Loss Functions:}
The combined output predictions for loss calculation can be summarised as:
\vspace{-0.5em}
\begin{equation}
    \mathcal{L} = {\mathcal{L}}_{encoder} + \sum_{N}{\mathcal{L}}_{decoder} \approx \sum_{N+1}{{(\mathcal{L}}_{ce} + \lambda {\mathcal{L}}_{smooth})}
    \vspace{-1em}
\end{equation}
Here, ${\mathcal{L}}_{encoder}$ represents the loss computed from the initial encoder predictions, followed by ${\mathcal{L}}_{decoder}$ arising from subsequent refinement stages in the $N-stage$ decoder. Each temporal stage involves the calculation of cross-entropy (${\mathcal{L}}_{ce}$) and a weighted smooth loss (${\mathcal{L}}_{smooth}$). The utilization of cross-entropy assists in capturing the dissimilarity between predicted and ground truth labels, while the weighted smooth loss helps mitigate over-segmentation, ensuring a more balanced and nuanced learning process.

\vspace{-0.5em}
\section{Experiments} 
\vspace{-0.5em}
\subsection{Dataset} \label{dataset}
\subsubsection{Cholec80}: The Cholec80 dataset~\cite{twinanda2016endonet} is a publicly accessible collection focusing on cholecystectomy surgery, encompassing more than 80 high-resolution videos conducted by 13 surgeons. Video frames come in dimensions of either $1920 \times 1080$ or $854 \times 480$ pixels, with an average runtime of 38-39 minutes at 25 frames per second (fps). Surgeons manually annotated the dataset, providing details on both the surgical phase and the surgical tools present in each frame. The surgical procedure is categorized into seven phases: Preparation, Calot triangle dissection, Clipping and cutting, Gallbladder dissection, Gallbladder packaging, Cleaning and coagulation, and Gallbladder retraction. In line with previous studies~\cite{gao2021trans,jin2017sv,twinanda2016endonet,yi2019hard,yi2021asformer,yi2022not}, the dataset is divided, with the initial 40 videos allocated for training and the remaining 40 utilized for testing.

\vspace{-1em}
\subsubsection{UroSlice}: We compiled the UroSlice dataset, comprising 39 videos documenting right and left partial and radical nephrectomy surgeries performed by two expert surgeons. A radical nephrectomy involves the comprehensive removal of the kidney, perinephric fat, and Gerota's fascia, while a partial nephrectomy entails surgically removing a renal tumor along with a rim of normal renal parenchyma, followed by reconstructing the defect using sutures. The surgeries utilized the `Da Vinci Xi' surgical robot developed by Intuitive Surgical Incorporated~\cite{carpentier1999computer}. Video recordings, with a resolution of $720 \times 480$ pixels, were captured on an SDC3 HD information management system by Stryker, having an average runtime of 100 minutes at 30 fps. Due to resource constraints, each video is downsampled to \textit{1 fps}. The dataset is meticulously labeled with phase annotations under expert supervision. Key phases, defined by a senior doctor in the partnering hospital, along with their durations, are documented in Table \ref{tab:uro_dataset}. For training and testing, we partitioned the dataset into 29 and 10 videos, respectively.

\vspace{-1em}
\begin{table}[!h]
\caption{List of Surgical Phases in the UroSlice dataset, including the mean $\pm$ std of the phases in minutes.}
\label{tab:uro_dataset}
\centering
\begin{tabularx}{0.9\textwidth}{>{\centering\arraybackslash}p{1cm} | >{\centering\arraybackslash}p{7cm} | >{\centering\arraybackslash}p{2.5cm}}
\hline
ID  & \multicolumn{1}{c|}{Task}                   & Duration (mins)   \\ \hline \hline
P1  & Reflect colon medially                      & 5.78 $\pm$ 3.84   \\
P2  & Dissect   renal vein                        & 2.29 $\pm$ 1.93   \\
P3  & Dissect   renal artery                      & 2.52 $\pm$ 1.61   \\
P4  & Ligate   hilum (radical)                    & 2.09 $\pm$ 2.38   \\
P5  & Identify and expose tumor (partial)         & 30.26 $\pm$ 13.84 \\
P6  & Excise   tumor (partial)                    & 8.35 $\pm$ 3.58   \\
P7  & Renorrhaphy   (partial)                     & 10.58 $\pm$ 5.59  \\
P8  & Ligate   ureter (radical)                   & 1.02 $\pm$ 0.63   \\
P9  & Lower pole and lateral dissection (radical) & 7.33 $\pm$ 5.44   \\
P10 & Put   tumor/kidney in bag                   & 1.40 $\pm$ 1.87   \\
P11 & Miscellaneous   tasks                       & 25.18 $\pm$ 10.33 \\\hline
\end{tabularx}
\end{table}

\vspace{-0.5cm}
\subsection{Experimental Setup}
The proposed architecture is implemented in PyTorch, utilizing a ResNet50~\cite{he2016deep_resnet} backbone for feature extraction. Optimization employs an Adam optimizer with a learning rate of $5 \times 10^{-4}$, weight decay of $1 \times 10^{-5}$, and cross-entropy loss during training. The `ViTALS' model comprises one encoder and three decoder modules, each consisting of $L=10$ blocks with hierarchical dilated convolution and attention layers, set at dimensions of $64$. The output is passed through a Fully Connected (FC) fusion head for final predictions. A sampling strategy evenly selects frames from all tasks during training to address the dominance of a few phases. The `ViTALS' model undergoes 150 epochs of training using the Adam Optimizer, with a learning rate of $5 \times 10^{-4}$ and a dropout rate of 0.3, ensuring effective feature extraction, hierarchical refinement, and robust training across diverse datasets.

\subsection{Results}
To assess our model's performance, we conducted a comparative analysis against state-of-the-art (SOTA) approaches for surgical phase recognition. We evaluated both video-level metrics~\cite{carreira2017quo,tao2013surgical} and sequence-level metrics~\cite{kuehne2016end}. Video-level metrics, including Accuracy (AC), assess frame predictions, while sequence-level metrics, such as Precision (PR), Recall (RE), and Jaccard (JA), evaluate phases/segments. Table~\ref{tab:result} provides a quantitative comparison between our model and SOTA methods on the Cholec80 and UroSlice datasets. Trans-SVNet~\cite{gao2021trans} and ASFormer~\cite{yi2021asformer} were implemented using the provided code, while other networks were extracted from their original works. Our model demonstrates a significant 1.4\% improvement in accuracy compared to other methods. Notably, it exhibits a lower standard deviation of 2.8\% when contrasted with Trans-SVNet, AVT, and MTRCNet-CL.

\vspace{-0.5em}
\begin{table}[!h]
\caption{Comparison of state-of-the-art Accuracy, Precision, Recall, and Jaccard scores (mean $\pm$ std \%) on Cholec80 and UroSlice datasets. \textbf{Bold} values denote the best results. `\#' denotes methods based on multi-task learning that require extra tool labels.}
\label{tab:result}
\resizebox{\textwidth}{!}{%
\begin{tabular}{p{1.7cm} p{3.4cm} p{2cm} p{2cm} p{2cm} p{2cm}}
\hline
\multirow{2}{*}{Dataset}   & \multirow{2}{*}{Method}                   & Video-level            & \multicolumn{3}{c}{Sequence-level}      \\ \cline{3-6} 
                           &                                           & Accuracy $\uparrow$    & Precision $\uparrow$     & Recall $\uparrow$        & Jaccard $\uparrow$     \\ \hline
\multirow{11}{*}{Cholec80} & EndoNet~\cite{twinanda2016endonet}$^{\#}$ & 81.1 $\pm$ 3.6         & 72.6 $\pm$ 13.8          & 79.6 $\pm$ 7.9           & $-$             \\
                           & EndoNet \texttt{+} LSTM~\cite{twinanda2017endolstm}$^{\#}$      & 86.4 $\pm$ 4.7  & 82.8 $\pm$ 8.1     & 81.2 $\pm$ 4.8           & $-$             \\
                           & MTRCNet-CL~\cite{jin2020multi}$^{\#}$     & 87.2 $\pm$ 5.6         & 83.9 $\pm$ 4.3           & 86.0 $\pm$ 5.7           & $-$             \\ \cline{2-6} 
                           & PhaseNet~\cite{twinanda2016single}        & 78.8 $\pm$ 4.7         & 71.3 $\pm$ 15.6          & 76.6 $\pm$ 16.6          & $-$             \\
                           & SV-RCNet~\cite{jin2017sv}                 & 84.3 $\pm$ 7.3         & 80.7 $\pm$ 6.8           & 82.5 $\pm$ 7.5           & $-$             \\
                           & OHFM~\cite{yi2019hard}                    & 86.3 $\pm$ 5.7         & $-$                      & $-$                      & 67.0 $\pm$ 12.3 \\
                           & TeCNO~\cite{czempiel2020tecno}            & 87.6 $\pm$ 7.8         & 85.5 $\pm$ 6.4           & 87.6 $\pm$ 6.7           & 75.1 $\pm$ 6.9  \\
                           & Trans-SVNet~\cite{gao2021trans}           & 89.1 $\pm$ 7.1         & \textbf{88.7 $\pm$ 5.0}  & \textbf{89.8 $\pm$ 7.4}  & \textbf{74.3 $\pm$ 6.6}  \\
                           & AVT~\cite{girdhar2021avt}                 & 86.7 $\pm$ 7.6         & 77.3 $\pm$ 4.8           & 82.1 $\pm$ 4.2           & 66.4 $\pm$ 6.4  \\
                           & ASFormer~\cite{yi2021asformer}            & 87.8 $\pm$ 2.8         & 79.2 $\pm$ 3.4           & 83.4 $\pm$ 5.3           & 76.8 $\pm$ 7.7  \\
                           & \textbf{ViTALS (Ours)}                                & \textbf{89.8 $\pm$ 4.1}& 82.1 $\pm$ 2.1           & 84.8 $\pm$ 4.8           & 74.8 $\pm$ 5.1  \\ \hline
\multirow{3}{*}{UroSlice}  & Trans-SVNet~\cite{gao2021trans}           & 62.2 $\pm$ 6.8         & 57.8 $\pm$ 8.1           & 61.1 $\pm$ 4.8           & $-$           \\
                           & ASFormer~\cite{yi2021asformer}            & 57.4 $\pm$ 4.2         & 52.8 $\pm$ 6.8           & 57.1 $\pm$ 8.1           & $-$           \\
                           & \textbf{ViTALS (Ours)}                                & \textbf{66.1 $\pm$ 2.1}& \textbf{59.2 $\pm$ 7.2}  & \textbf{63.1 $\pm$ 5.4}  & $-$           \\ \hline
\end{tabular}%
}
\end{table} 

\begin{figure}[!h]
\includegraphics[width=\textwidth]{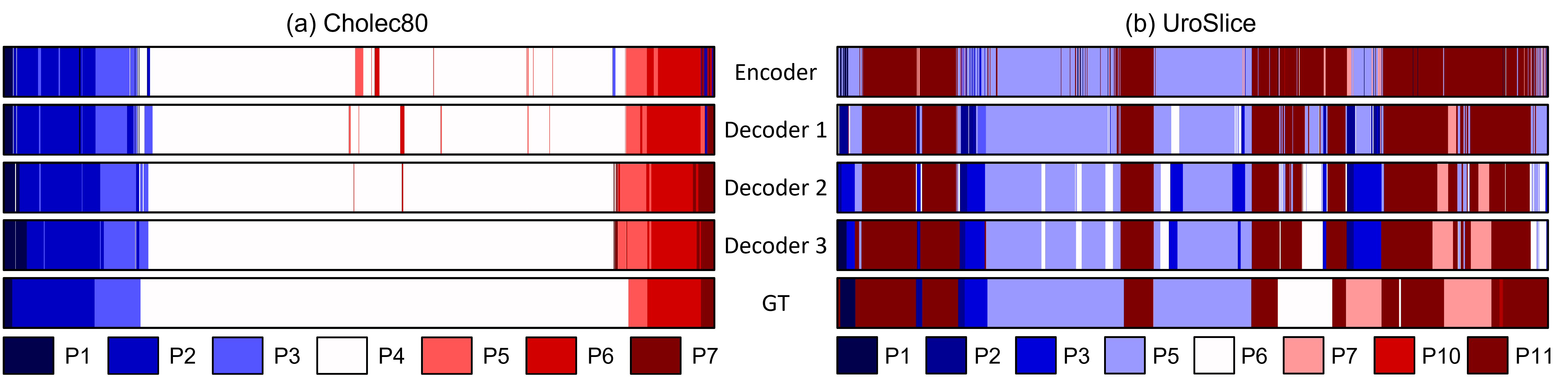}
\caption{Illustrations depicting the qualitative results of surgical phase recognition for (a) Cholec80 and (b) UroSlice datasets are presented. Each line represents the predicted output at different stages of the segmentation network, aligned with the corresponding ground truth labels (GT). Our `ViTALS' model demonstrates the refinement and produce comparable results to the ground truth predictions for both datasets.}  \label{fig:results}
\end{figure}

\vspace{-0.5em}
The UroSlice dataset is characterized by significant variations in the duration of its phases as evident in Figure \ref{fig:results}. The occurrences of these phases lack uniformity and systematic order, contributing to a dataset that presents challenges due to its unpredictable nature and substantial differences in temporal aspects. Under these conditions, our model surpasses SOTA models like Trans-SVNet and ASFormer by a margin of 5.9\% in accuracy and approximately 2\% in terms of precision and recall. This underscores the robustness of our model, particularly in handling smaller datasets with complex workflows. Overall, our model consistently outperformed others in both video-level and sequence-level evaluations, as indicated by the comprehensive evaluation results in Table \ref{tab:result}.

\subsection{Ablation study}
Feature extractor used in our network plays an important role during the phase action segmentation. We conduct an ablation study by varying the network used for feature extraction and report the GPU memory cost during training in Table \ref{tab:ablation}. Model 1 uses a shallow 3-layer convolution network to reduce the dimension of images $H (256) \times W (256) \times C (3) = 196608$ to $d = 2048$. Model 2 uses a patch embedding network used in vanilla ViT~\cite{dosovitskiy2020image_vit} to create feature vectors of dimension $d = 2048$. Our `ViTALS' model uses a ResNet50 as the backbone to extract the features as discussed in Section \ref{feature}.

\vspace{-0.5em}

\vspace{-0.8cm}
\begin{table}[!h]
\caption{Ablation study on Cholec80 and UroSlice dataset. (a) Comparison of various architectures used for feature extractor (FE) network (b) Effect of multi-stage decoder. }
\label{tab:ablation}
\centering
\begin{tabularx}{0.78\textwidth}{p{3cm} *{2}{>{\centering\arraybackslash}p{2cm}} >{\centering\arraybackslash}p{2.2cm}}
\hline
\multicolumn{1}{l}{\multirow{2}{*}{(a)     FE Network}} & \multicolumn{2}{c}{Accuracy} & \multicolumn{1}{c}{\multirow{2}{*}{GPU Memory}} \\ \cline{2-3}
                            & \multicolumn{1}{c}{Cholec80} & \multicolumn{1}{c}{UroSlice}          &    \\ \hline
Conv layer                                      & 58.3 $\pm$ 7.2 & 30.2 $\pm$ 5.4   & $\sim$ 402M        \\
Patch Embedding                                 & 81.6 $\pm$ 3.6 & 42.5 $\pm$ 6.5   & $\sim$ 125M        \\
ResNet50                                        & 89.8 $\pm$ 4.1 & 66.1 $\pm$ 2.1   & $\sim$ 27.7M       \\ \hline
\end{tabularx}
\centering
\begin{tabularx}{0.78\textwidth}{p{3cm} *{4}{>{\centering\arraybackslash}p{1.5cm}}}
\hline
\multirow{2}{*}{(b)   Segmentation} & \multicolumn{2}{c}{Accuracy} & \multicolumn{2}{c}{Precision} \\ \cline{2-5}
                              & Cholec80        & UroSlice        & Cholec80        & UroSlice         \\ \hline
Encoder only                  & 84.8            & 54.8       & 64.2            & 42.7        \\
Encoder \texttt{+} Decoders            & 89.8            & 66.1       & 82.1            & 61.2        \\ \hline
\end{tabularx}
\end{table}

\vspace{-1em}
\section{Conclusion}
In our research, we present the ViTALS model, which leverages advanced temporal-rich spatial feature modeling and introduces a multi-stage paradigm for temporal feature refinement. This approach effectively integrates local fine-grained details and global dependencies. Our model exhibits comparable performance to state-of-the-art models on the relatively straightforward Cholec80 dataset. Importantly, for more challenging datasets such as UroSlice, we observe a significant improvement, achieving a 5.9\% increase in accuracy and approximately 2\% enhancement in precision and recall. These results highlight the robustness of our model, particularly in effectively handling smaller datasets characterized by intricate workflows.

\section*{Acknowledgement}
This work was supported in part by the Center for Co-design of Cognitive Systems (CoCoSys), one of the seven centers in JUMP 2.0, a Semiconductor Research Corporation (SRC) program sponsored by DARPA, by the SRC, the National Science Foundation, Intel Corporation, the DoD Vannevar Bush Fellowship, and by the U.S. Army Research Laboratory.

%
%
%
\newpage
\bibliographystyle{splncs04}
\bibliography{references}
\end{document}